%% file: main.tex
\documentclass[9pt,conference]{IEEEtran}
\IEEEoverridecommandlockouts

\usepackage{flushend}
\usepackage{threeparttable}
\usepackage{cite}
\usepackage{amsmath,amssymb,amsfonts}
\usepackage{algorithmic}
\usepackage{graphicx}
\usepackage{textcomp}
\usepackage{xcolor}

\usepackage{enumitem}
\usepackage[utf8]{inputenc} 
\usepackage[T1]{fontenc}    
\usepackage{hyperref}       
\hypersetup{
    colorlinks=true,
    linkcolor=blue,
    citecolor=blue,
    filecolor=magenta,      
    urlcolor=cyan,
    pdftitle={Overleaf Example},
    pdfpagemode=FullScreen,
    }
    
\usepackage{url}            
\usepackage{booktabs}       
\usepackage{amsfonts}       
\usepackage{nicefrac}       
\usepackage{microtype}      
\usepackage{xcolor}         
\usepackage{graphicx}
\usepackage{booktabs} 
\usepackage{amsmath}
\usepackage{makecell}

\usepackage{url}  
\usepackage{multirow}
\usepackage{amsmath}
\usepackage{amsthm}
\usepackage{pifont}

\def\BibTeX{{\rm B\kern-.05em{\sc i\kern-.025em b}\kern-.08em
    T\kern-.1667em\lower.7ex\hbox{E}\kern-.125emX}}

\begin{document}

\title{On-Device Unsupervised Image Segmentation
}
\author{
\IEEEauthorblockN{
Junhuan Yang$^{1}$ \quad
Yi Sheng$^{2}$ \quad
Yuzhou Zhang$^{3}$ \quad
Weiwen Jiang$^{2}$ \quad
Lei Yang$^{1}$ \quad
}
\IEEEauthorblockA{
\normalsize
$^{1}$ Information Sciences and Technology Department, George Mason University\\
$^{2}$ Department of Electrical and Computer Engineering, George Mason University\\
$^{3}$ Khoury College of Computer Sciences, Northeastern University
}
}

\maketitle

\begin{abstract}
Along with the breakthrough of convolutional neural networks, in particular encoder-decoder and U-Net, learning-based segmentation has emerged in many research works.
Most of them are based on supervised learning, requiring plenty of annotated data; however, to support segmentation, a label for each pixel is required, which is obviously expensive.
As a result, the issue of lacking annotated segmentation data commonly exists.
Continuous learning is a promising way to deal with this issue; however, it still has high demands on human labor for annotation. What's more, privacy is highly required in segmentation data for real-world applications, which further calls for on-device learning.
In this paper, we aim to resolve the above issue in an alternative way: Instead of supervised segmentation, we propose to develop efficient unsupervised segmentation which can be executed on edge devices without annotated data.
Based on our observation that segmentation can obtain high performance when pixels are mapped to a high-dimension space using their position and color information, we for the first time bring brain-inspired hyperdimensional computing (HDC) to the segmentation task. 
We build the HDC-based unsupervised segmentation framework, namely ``SegHDC''.
In SegHDC, we devise a novel encoding approach, which follows the Manhattan distance.
A clustering algorithm is further developed on top of the encoded high-dimension vectors to obtain segmentation results.
Experimental results show that SegHDC can significantly surpass neural network-based unsupervised segmentation.
On a standard segmentation dataset, DSB2018, SegHDC can achieve a 28.0\% improvement in Intersection over Union (IoU) score; meanwhile, it achieves over 300$\times$ speedup on Raspberry PI.
What's more, for a larger size image in the BBBC005 dataset, the existing approach cannot be accommodated to Raspberry PI due to out of memory; on the other hand, SegHDC can obtain segmentation results within 3 minutes while achieving a 0.9587 IoU score.

\end{abstract}

\input{I_Introduction.tex}
\input{II_Motivation.tex}

\input{III_Framework.tex}

\input{IV_Exp}

\input{V_Conclusion}

{\tiny
\bibliography{paper.bib}
\bibliographystyle{IEEEtran}
}
\end{document}

%% file: I_Introduction.tex
\section{Introduction} \label{sec:Intro}

Segmentation is a fundamental task in a lot of applications, such as 
shadow detection
and medical imaging \cite{liao2021shadow, wu2021federated, yang2020co-exploration}. 
With the  fast development of artificial intelligence (AI), targeting the segmentation tasks, there emerge both manually developed deep neural network architectures (e.g., encoder and decoder \cite{noh2015learning, badrinarayanan2017segnet} and U-Net \cite{ronneberger2015u}) as well as automated generated architectures (e.g., Auto-DeepLab \cite{liu2019auto}).
These deep learning models have demonstrated superior performance on benchmarking datasets against segmentation in traditional imaging processing.
However, when it comes to real-world applications, the lack of annotated data becomes a critical issue: on the one hand, the annotation demands human labor to label data which is costly, not mention to the segmentation tasks that require labels for each pixel;
on the other hand, the lack of training data will drastically degrade segmentation performance.
In consequence, how to perform segmentation without plenty of annotated data is highly desired.

With such a need, unsupervised learning seems to be a natural answer, since it can discover useful patterns in data without annotation \cite{celebi2016unsupervised}. 
It seems straightforward to apply unsupervised learning to segmentation, however, the segmentation task itself and the demands from the application bring new challenges.
First, unlike unsupervised classification (a.k.a, clustering), segmentation requires the process of features for every pixel to figure out which ones are compact enough to form distinct clusters.
As such, unsupervised segmentation can require a deeper neural architecture, which makes the model much larger.
Second, real-world segmentation applications commonly have data privacy demands (e.g., medical imaging) and real-time requirements (e.g., autonomous vehicles).
The ideal solution to address data privacy is to process data on-device~\cite{yang2020co, jiang2019accuracy, jiang2020hardware, jiang2019achieving, zhan2021accelerating, zhan2018energy}; however, the edge devices have limited computation resources and they are facing a large model size, both of which conflict with the real-time requirement.

To address the above challenges, we are rethinking what is the best computing model to extract features for segmentation tasks.
For a long while, due to the superior performance of neural networks, they are typically adopted in unsupervised segmentation \cite{xia2017w, kim2020unsupervised}.
However, as stated, it requires a large model size, which may easily exceed the capacity of edge devices.
In segmentation, the pixel position and color are the most important information, and we find that it can achieve high performance if we smartly map the pixels into a high-dimensional space, according to the position and color information.

With such motivation, in this paper, we for the first time bring a recent emerging computing model, i.e., brain-inspired hyperdimensional computing (HDC) \cite{ge2020classification}, into image segmentation.
HDC has shown its superiority in robustness, scalability, and high energy efficiency for classification tasks \cite{zhang2021assessing, zhang2022scalehd, yangautomated, yang2022hardware, zhang2022energy}.
Fundamentally, HDC encodes data into a high-dimensional space using a hypervector (HV), and learns features in that space \cite{sheng2023toward}.
It seems that HDC is naturally suitable for segmentation; however, this characteristic of HDC has not been well developed and utilized. 
To bring HDC to segmentation, the first task is to figure out how to encode pixels such that the encoded pixels can 
precisely describe the disparity.
What's more, since the feature extracted by HDC is represented in high-dimension; it is challenging to efficiently perform clustering on high-dimension vectors.

To address these challenges, we propose a novel framework, namely SegHDC.
Through a fundamental analysis of the characteristics of HDC and image segmentation, we first propose a brand-new HDC encoding approach to encode both position and color information into high dimensional space. Then, a revised K-Means algorithm has been devised to cluster encoded HVs and label corresponding pixels.

The main contributions of this paper are as follows.
\begin{itemize}
    \item To the best of our knowledge, this is the first work to perform on-device unsupervised image segmentation.
    \item We innovatively apply hyperdimensional computing for image segmentation tasks with a brand-new design for encoding images. 
    \item Evaluation results verify the  effectiveness of SegHDC, which surpasses  state-of-the-art unsupervised CNN-based algorithm.
\end{itemize} 

We have carried out a set of experiments on 3 commonly used segmentation datasets to evaluate the effectiveness of our proposed SegHDC. 
Experiment results on the DSB2018 dataset
show the efficiency of SegHDC, specifically by outperforming the CNN-based baseline with 28.0\% improvement in Intersection over Union (IoU) score; meanwhile, achieving over 300$\times$ speedup on the edge device.
On the BBBC005 dataset with a larger size of images, SegHDC achieves an IoU score of 0.9414 which is 25.7\% higher than the score obtained by the baseline. SegHDC also obtains a 0.9587 IoU score for a sample image in the BBBC005 dataset with a latency of only around 178 seconds, while the existing CNN-based unsupervised segmentation approach cannot predict this image due to the computing resources limitation. On the other dataset, MoNuSeg, SegHDC also gains an improvement of 8.27\% compared with the baseline method.

The remainder of the paper is as follows: Section~\ref{sec:rel} presents
the related work and motivation. Section~\ref{sec:method} demonstrates our proposed SegHDC framework. Experimental results and conclusion are in Section~\ref{sec:exp} and Section~\ref{sec:conclusion} respectively.






%% file: II_Motivation.tex
%




\vspace{3pt}
\section{Related Work and Motivation}\label{sec:rel}
\vspace{3pt}

This section will first discuss the need for unsupervised segmentation and related work. 
Then, we provide our observation which motivates to use hyperdimensional computing for segmentation.

\noindent\textbf{Need and challenge: Unsupervised and on-device learning for image segmentation is highly demanded.} 

Image segmentation is a typical task in machine learning, and supervised learning has a high cost of labeling.
Unlike the classification task that one image needs only one label \cite{sheng2022larger, hu2022design, jiang2020standing, zhan2021improving}, the labeling of the segmentation dataset requires the assignment of a class to each pixel in an image.
What's worse, since segmentation is largely required in domain-specific applications, like medical imaging, the labeling task commonly calls for domain expertise.
For example, only good Computed Tomography (CT) doctors can distinguish if there is accurate lesion exists and where the accurate lesion is.
The involvement of doctors to do the labeling work is obviously too costly.
To overcome the high cost of labeling for segmentation tasks, unsupervised learning is highly demanded, which does not need labels of data to perform the segmentation tasks.
What's more, unsupervised segmentation can be applied to perform the automated annotation.

Although promising, unsupervised segmentation commonly requires a longer time over the inference of supervised learning.
What's more, when unsupervised segmentation is applied to real-world applications, with the consideration of data security, in-situ and real-time processing on edge devices is typically required.
Besides, the high cost of unsupervised learning and the limited computing resources on edge devices make the problem more challenging.

\vspace{3pt}
\noindent\textbf{Related work: Supervised segmentation has been widely studied, but unsupervised segmentation is still in its infancy.} \vspace{3pt}

The very first deep learning-based image segmentation was proposed in \cite{long2015fully}, which used a fully convolutional network (FCN) to perform segmentation. Another popular method used encoder-decoder
architecture segment the images \cite{noh2015learning, badrinarayanan2017segnet}. Inspired by the FCN and encoder-decoder architecture, U-Net \cite{ronneberger2015u} and V-Net \cite{milletari2016v} were proposed, which largely prompt the segmentation performance. 
Unsupervised segmentation becomes active recently.
Authors in \cite{moriya2019unsupervised} presented a novel unsupervised segmentation method for the 3-D
microstructure of lung cancer specimens in micro-computed tomography 
images. A CNN-based unsupervised image segmentation method was proposed in \cite{kim2020unsupervised}, and a clustering function was used after the CNN to cluster the pixels. Generative adversarial networks (GANs) were also used for unsupervised image segmentation \cite{abdal2021labels4free}. The authors utilized the features generated by a GAN and trained the segmentation networks.

Recently, there emerge research efforts in on-device learning for better deploying ML models on edge devices.
Some traditional ML 
models, like SVMs, can be directly accommodated on edge devices \cite{dhar2021survey}. In \cite{zhou2021octo}, the authors 
employ the ``int8'' quantization in both forward and backward passes over a deep model to enable on-device learning. Neural architecture search (NAS) is another solution to find tiny ML models and on-device training, like MCUNets \cite{lin2022device}.  

However, there is still a missing link between unsupervised segmentation and on-device learning.
Instead of repeatedly designing algorithms based on convolutional neural networks, we believe new innovations are needed to fill such a gap.

\vspace{3pt}
\noindent\textbf{Observation: Vectorizing pixels in a high-dimension space can map pixels in a similar area.} 
\vspace{3pt}

\begin{figure}[t]
\begin{center}
\includegraphics[width=3in]{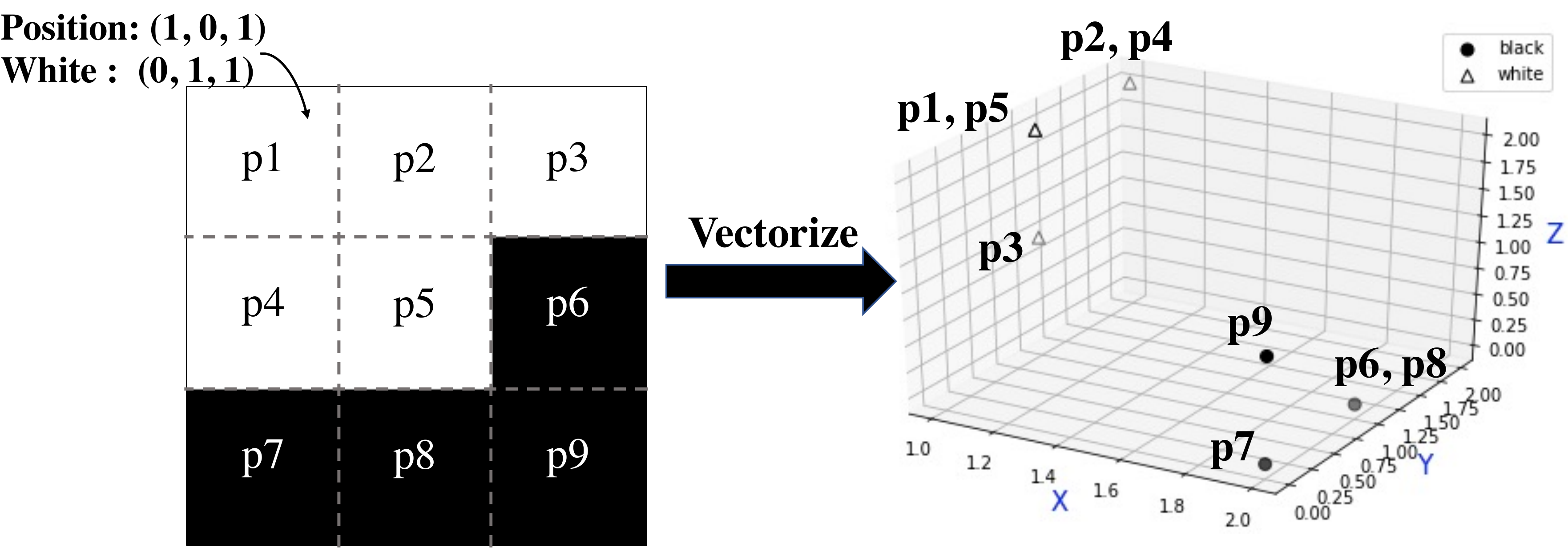}
\vskip -0.1in
\caption{Vectorized pixels with the same color have a short distance in space.}
\label{fig:mot1}
\end{center}
\vskip -0.1in
\end{figure}

The previous segmentation works mainly employ convolutional neural networks (CNNs) to extract the feature of the pixels, which brings high computation costs and can easily become the performance bottleneck on edge devices.
In this paper, we aim to simplify the feature extraction process for segmentation.
Specifically, each pixel has two fundamental information: (1) position, and (2) color. The question is whether we can perform segmentation by only exploring the spatial and color correlation of pixels.

To this end, we perform vectorization of pixel position and color and map each pixel in a high-dimension space.
To enable the visualization, we limit the vector dimension to 3, as the example shown in Figure \ref{fig:mot1}.
In this example, we generate a $3\times 3$ binary image. 
We randomly assign a binary vector to each row and column, and use $XOR$ to associate the vectors between a row and a column to generate the vector of a position. 
We also randomly generate a vector for each color.
For example, the first pixel has the vector of $(1,0,1)$ for position and $(0,1,1)$ for the white color.
The second pixel has a different position vector $(1,1,1)$ and the same vector for color.
After we obtain these vectors, we sum up the pair of vectors at the same position to map the pixel to the 3-dimension space.
The right-hand in Figure \ref{fig:mot1} shows the distribution of all these 9 pixels in the Cartesian coordinate system. 
We can easily observe that all the white pixels are mapped to a small area while the black pixels are mapped to another distinct area.



\vspace{3pt}
\noindent\textbf{Motivation: Using hyperdimensional computing (HDC) for unsupervised segmentation.} 
\vspace{3pt}

The above example gives us the hint that mapping image pixels to high-dimension space can be an effective way to perform the segmentation.
We are inspired and novelly involve hyperdimensional computing for segmentation process and apply it to more complicated images, to better represent pixels using vectors with much higher dimensions (e.g., 10,000) than the 3-dimension vectors used in the above example.

Unlike the traditional ML algorithms to directly process data, 
HDC will encode data into a high-dimensional space, where the data are represented by high-dimensional and pseudo-orthogonal hypervectors (HVs) \cite{ge2020classification}.
For an HV with dimension $d$, we can denote it as $
    \overrightarrow{H} \  = \ (\overrightarrow{e_1}, \ \overrightarrow{e_2}, \ ..., \ \overrightarrow{e_i}, \ ..., \ \overrightarrow{e_d})$,
where $\overrightarrow{e_i}$ represents the $i-th$ element in HV $\overrightarrow{H}$. 
In this way, data can be encoded and features can be extracted in the high-dimensional space.
Existing works have shown that HDC can work well on image classification \cite{liang2022distrihd}.

To the best of our knowledge, there is no work to use HDC for segmentation tasks. It seems straightforward to extend HDC designed for classification to perform segmentation tasks.
However, most of the existing classification approaches rely on the randomly generated vectors to represent pixels, while the relative relationship between different positions and between different colors is a key to extracting features in segmentation, which is not studied.
As explored in our work, together with experimental results, directly applying the existing approach cannot perform a good segmentation.
Therefore, a more dedicated design of position and color embedding is needed. With such vision, we have proposed a holistic framework to complete these tasks with details in Section \ref{sec:method}.

%% file: III_Framework.tex
\section{SegHDC Framework}\label{sec:method}

\begin{figure}[t]
\begin{center}
\includegraphics[width=\columnwidth]{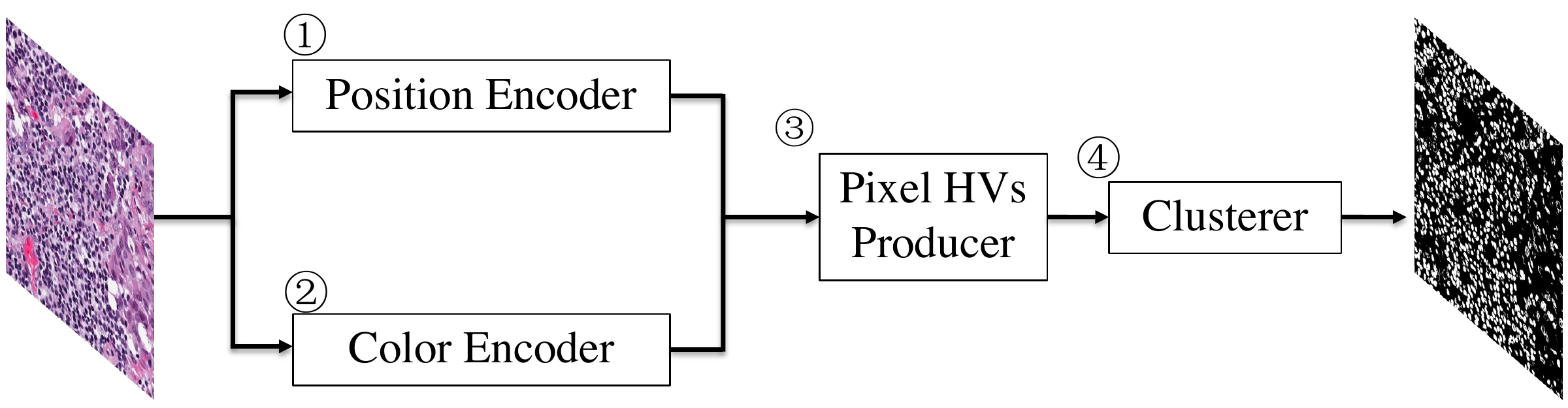}
\vskip -0.1in
\caption{Overview of the SegHDC framework.}
\label{fig:frame}
\end{center}
\vskip -0.1in
\end{figure}

In this section, we will formally introduce the SegHDC framework. As shown in Figure \ref{fig:frame}, SegHDC is mainly composed of 4 components: \textcircled{\raisebox{-1pt}{1}} 
position encoder module,
\textcircled{\raisebox{-1pt}{2}} color encoder module,
\textcircled{\raisebox{-1pt}{3}} pixel HV producer, 
and \textcircled{\raisebox{-1pt}{4}} clusterer. 
In the rest of this section, each component will be introduced in detail. 

\noindent\textbf{\textcircled{\raisebox{-1pt}{1}} Position Encoder}




Spatial information is one of the key pieces of information in one image, which is particularly important for segmentation. We need to clearly represent and measure both the \textit{relationship} and the \textit{difference} of pixels in different positions.
Manhattan distance is a typical distance metric used to measure different positions in the 2-dimensional space (row and column), which will also be used in this paper to guide the position encoding.
The main concept of Manhattan distance is that the distance between two points is the sum of the absolute differences of their Cartesian coordinates \cite{ManhattanDist}, i.e., $L1$ norm. In a plane with point $p_1$ at $(x1, y1)$ and point $p_2$ at $(x2, y2)$, their Manhattan distance is calculated as $d_1(p_1, p_2) = \left | x_1 - x_2 \right | + \left | y_1 - y_2 \right |$.
When we extend these two points $p$ and $q$ in a $n$ dimension space, the Manhattan distance can be described as Equ \ref{equ:Man2}, where, $p = (p_1, p_2, ..., p_n), q = (q_1, q_2, ..., q_n)$.
\vskip -0.1in
\begin{equation}
\small
\begin{aligned}
    d_1(p, q) = \sum_{i = 1}^{n}\left | p_i - q_i \right | 
    \label{equ:Man2}
\end{aligned}
\end{equation}
\vskip -0.1in
Given any three pixels in a single channel image, $p_1$ at $i-th$ row and $j-th$ column, $p_2$ at  $m_0-th$ row and  $n_0-th$ column and $p_3$ at  $m_1-th$ row and  $n_1-th$ column, we let the points meet Equ \ref{equ:Man4}.
\begin{equation}
\small
\begin{aligned}
d_1(p_1, p_2) =  d_1(p_1, p_3), \ iff \  m_0 + n_0 = m_1 + n_1\\
for \ \forall p_1 = (i,j), \forall p2 = (m_0, n_0), \forall p3 = (m_1, n_1)
    \label{equ:Man4}
\end{aligned}
\end{equation}
Inspired by randomly flipped elements in HVs in the traditional encoding method, we are going to use the element flip to represent the distance. To this end, we need to control the row and column at the same time and treat the image matrix (single channel) as in two dimension space instead of one dimension in the traditional encoding method. 
Take the rows HVs as the example, we first randomly generate a binary HV (all elements are ``0'' and ``1'' ) with a high dimensionality $d$, e.g., 10,000. 
The second step is to flip the next $x_{row}$ (shown in Equ \ref{equ:uRow}, where $N_{row}$ and $N_{column}$ means the number of rows/columns) elements. The process will be continued until the last row.
A similar process will be conducted to columns, while the number of elements that need to be flipped is defined as $x_{col}$, shown in Equ \ref{equ:uRow}.
\begin{equation}
\small
    x_{row} = \left \lfloor \frac{d}{N_{row}} \right \rfloor, \ x_{col} = \left \lfloor \frac{d}{N_{column}} \right \rfloor 
    \label{equ:uRow}
\end{equation}
The most important step is to use an HV to represent the pixel at $i-th$ row and $j-th$ column. Classical HDC uses element-wise XOR or 
multiplication to associate two HVs. Element-wise multiplication can not keep the distance, since any 0 can diminish the distance, and we can not control where the 0 should be. Element-wise XOR is naturally suitable for this operation. Thus we get the position HV to represent the pixel at $i-th$ row and $j-th$ column as $y_{(i,j)} = r_i \oplus c_j$.
Thus, we can use the position HVs to describe the Manhattan distance, and we can have Equ \ref{equ:Man5}.
\begin{equation}
\small
\begin{aligned}
d_1(r_i \oplus c_j, r_{i+m_0} \oplus c_{j+n_0}) =  d_1(r_i \oplus c_j, r_{i+m_1} \oplus c_{j+n_1}),\\
for \ \ \forall i,j, \forall m_0, n_0, \forall m_1, n_1,\  iff \ \  m_0 + n_0 = m_1 + n_1
    \label{equ:Man5}
\end{aligned}
\end{equation}

\begin{figure}[t]
\begin{center}
\includegraphics[width=0.8\columnwidth]{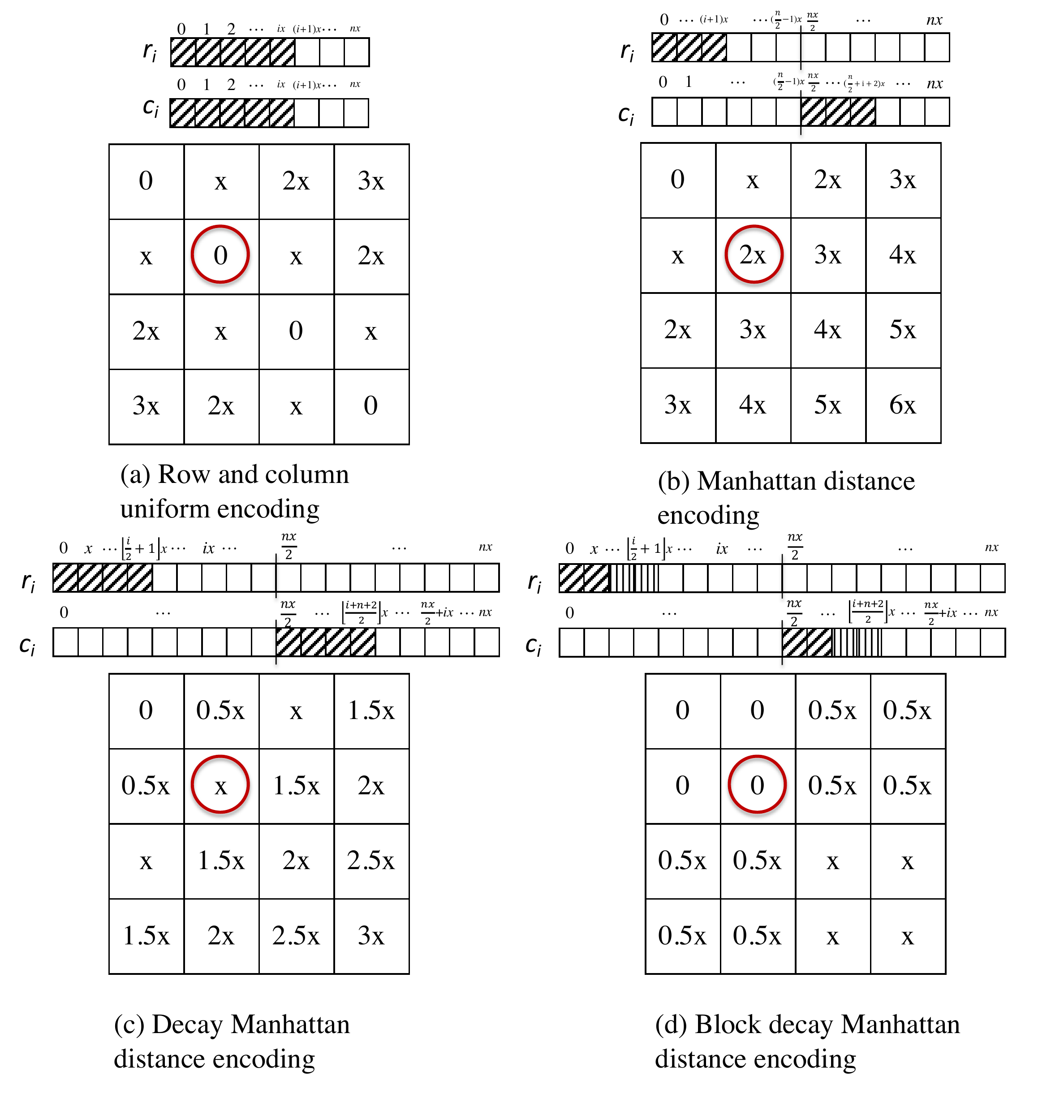}
\vskip -0.1in
\caption{Distance between HV at 0-th row, 0-th column and others, when $\alpha$ and $\beta$ set as 0.5 and 2}
\label{fig:distance}
\end{center}
\vskip -0.1in
\end{figure}


The distance of the above encoding is shown in Figure \ref{fig:distance} (a). All the distances shown in the position are the distance between the position HV at $p_{(0,0)}$ ($0-th$ row and $0-th$ column) and the position HV at $p_{(i,i)}$ ($i-th$ row and $i-th$ column). 
For example, the distance between $p_{(0,0)}$ and $p_{(1,1)}$ is 0, shown in the 
position at the intersection of the second row and second column(in the red circle). The $x$ means the $x_{row}$ and $x_{column}$ defined in Euq \ref{equ:uRow}. 
To simplify the question, we suppose $x_{row}$ = $x_{col}$ = $x$ here.
The zeroth row and zeroth column seem to meet the Manhattan distance, while other parts do not. For example, the distance between $p_{(0,0)}$ and $p_{(1,1)}$ (in the red circle) should be $2x$ instead of $0$.
This is because, for the pixel at $i-th$ row and $i-th$ column, the $r_i$ and $c_i$ flip the element on the same sites (shown in the top part of Figure \ref{fig:distance} (a)). Thus $r_1$ $\oplus$ $c_1$ = $r_i$ $\oplus$ $c_i$, and all the distances between $p_{(1,1)}$ and $p_{(i,i)}$ is 0. Similar distance diminishing occurs in other parts.
So, we need to let the HVs of rows and columns flip different sites. Specifically, row HV can change the first half of row HV while column HV can change the second half of column HV. Thus the changes in row HV  will not affect the changes in column HV, and vice versa. The distance of this encoding is shown in Figure \ref{fig:distance}  (b), and it meets Equ \ref{equ:Man5} now. 
We call this encoding method Manhattan distance encoding. 
However, all the distances add $x$ from the closer one and it can not describe a smaller distance. Thus we bring a new hyperparameter $\alpha$ to 
describe the ratio of the half dimension which needs to be changed and thus control the flip unit. This encoding is called decay Manhattan distance encoding, and the distances when $\alpha$ set as 0.5 is shown in Figure \ref{fig:distance} (c). 
And thus Equ \ref{equ:uRow} 
changes to Equ \ref{equ:uRow2}.
\begin{equation}
    x_{row} = \left \lfloor  \frac{\alpha \times d}{2 \times N_{rows}}\right \rfloor,\ x_{col} = \left \lfloor \frac{\alpha \times d}{2 \times N_{col}} \right \rfloor
    \label{equ:uRow2}
\end{equation}
It can also meet Equ \ref{equ:Man5} and the smaller distance can be described. 
Another issue is that continuous near pixels are much more likely to be annotated as the same label. So it seems that the position HVs in a small area should not change.
Thus, we bring the second hyperparameter $\beta$ to let $\beta$ rows and $\beta$ columns as a block, and Manhattan distance is computed based on these blocks. 
We call this encoding method block decay Manhattan distance encoding.
Figure \ref{fig:distance} (d) shows the distance when $\beta$ is set as 2. 
And all the distance between HVs of two positions should meet the new block Manhattan distance, as shown in Equ \ref{equ:Man6}. 
\begin{equation}
\small
\begin{aligned}
d_1(b_{r_i} \oplus b_{c_j}, b_{r_{i+m_0}} \oplus b_{c_{j+n_0}}) =  d_1(b_{r_i} \oplus b_{c_j}, b_{r_{i+m_1}} \oplus b_{c_{j+n_1}}),\\
iff \ \  m_0 + n_0 = m_1 + n_1, \ for \ \ \forall i,j, \forall m_0, n_0, \forall m_1, n_1
    \label{equ:Man6}
\end{aligned}
\end{equation}




\noindent\textbf{\textcircled{\raisebox{-1pt}{2}} Color Encoder}

The color value is in a one-dimension space, and it varies from 0 to 255. However, we have three channels in most images. These three channels have the same position but possibly different color values. 
To simplify the question, we consider a single channel at first. 
Classical HDC randomly generates 256 HVs to represent the 256 values of color, or randomly flips several elements based on the previous HV.
While random means 
color values with a greater difference
can be assigned to more similar HVs, which does not make sense. 
Similar to position encoding, we need to encode the color value according to the Manhattan  distance defined in Equ \ref{equ:Man2}. 
The Manhattan distance of two HVs is the number of different elements in the HVs, thus we can also use the summation of the element-wise XOR to obtain it. Similar to the position HVs, we can flip the number of continuous sites of HVs to add the Manhattan
distance to the HVs. Since the value variance is fixed from 0 to 255, we can define the unit length per flip as $u_c$ as $u_c = \left \lfloor \frac{d}{256} \right \rfloor $.

The Manhattan
distance of two color values $a$ and $b$ can be described as $ d_1(a,b) = \left \lfloor \frac{ \left | a - b \right | }{256} \right \rfloor $
while the largest Manhattan distance, $\frac{255}{256} \times u_c $, is between the HVs corresponding to $0$ and $255$. Thus, we obtain the 256 HVs of 256 single-channel color values. 



\begin{figure}[t]
\begin{center}
\includegraphics[width=\columnwidth]{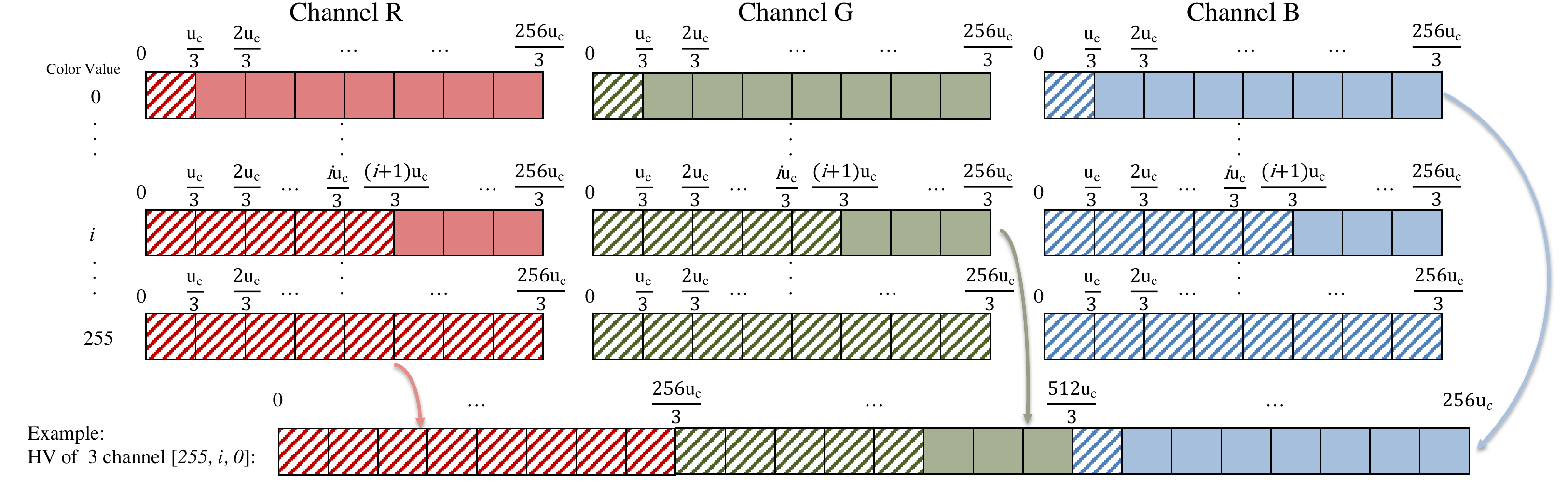}
\vskip -0.1in
\caption{3-channel color encoding under the guidance of Manhattan distance.
The block with diagonal stripes means the elements in this block are flipped.}
\label{fig:3channel}
\end{center}
\vskip -0.1in
\end{figure}

To encode the 3-channels color value at one position actually encodes three different values into one HV with dimension $d$. From classical HDC, we may use element-wise $XOR$ or $multiplication$ to associate two HVs and then three. In fact, the difficulty is how to keep the 
Manhattan distance. However, neither element-wise $XOR$ nor element-wise $multiplication$ can keep the Hamming distance. Element-wise $multiplication$ can diminish the distance when there is a $0$ at the same site of any one of three HVs, while element-wise $XOR$ only counts number of 1s.
So we need a new way to encode the 3-channels color values.
To contain all the information of the three channels and keep the
Manhattan distance, we consider reducing the HV dimension of each channel from $d$ to $\frac{d}{3}$. Thus we can get a $3 \times 256 \times \frac{d}{3}$ HVs to represent the color values in 3 channels, and each channel obtain $256 \times \frac{d}{3}$ HVs. At last, when the three values in the three-channel are obtained, the three corresponding HVs will be concatenated together to produce a new HV to represent the color value of this pixel. As shown in Figure \ref{fig:3channel}, if the color values of a pixel are [255, i, 0], the HV to represent its color should consist of three parts:
(1) the first $\frac{d}{3}$ elements come from the $256th$ HV (value 255) of the first channel, (2) the second $\frac{d}{3}$ elements come from the $(i+1)-th$ HV (value $i$) of the second channel, and (3) the rest elements come from the $1st$ HV (value $0$) of the third channel.



\noindent\textbf{\textcircled{\raisebox{-1pt}{3}} Pixel HV Producer}
\begin{figure}[t]
\begin{center}
\includegraphics[width=\columnwidth]{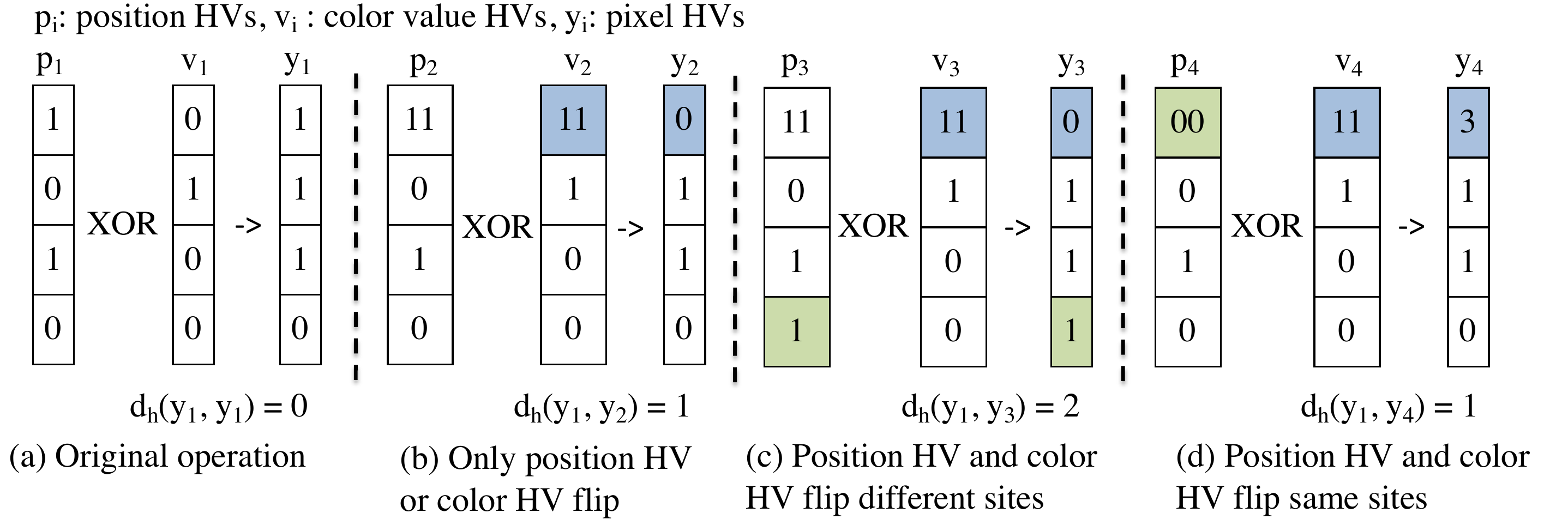}
\vskip -0.1in
\caption{Different situations when producing pixel HV, and relative Manhattan distance changes when $\gamma = 2$ is applied to color HV. 
}
\label{fig:pixelHV}
\end{center}
\vskip -0.1in
\end{figure}

After the position and color values of the pixel are encoded to HVs, we will produce the pixel HVs based on these two types of HVs. The only issue that needs to be concerned about is how can we 
keep the Manhattan distance in position HVs and
color HVs. 
We use Hamming distance $d_h$ as guidance here. It describes the number of positions with the different symbols in two equal-length strings \cite{waggener1995pulse}.
To help adjust the composition ratio of position HV and color HV, we also bring a new hyperparameter $\gamma$. 
$\gamma$ makes the flipped element longer to the $\gamma$ bit filling in the flipped value, which can impact on either position HV or color HV. For example, when $\gamma = 2$, and $0$ can flip to $1$ and then change as long as to be $11$.
Figure \ref{fig:pixelHV} shows the different situations when producing pixel HV, and the relative Hamming distance changes when $\gamma = 2$ is applied to color HV $v_i$. 
In Figure \ref{fig:pixelHV} (b), color HV $v_2$ flips one element, thus the resulting pixel HV $y_2$ flips the same site. 
The Hamming distance is 1 between the new pixel HV $y_2$ and the original pixel HV $y_1$. 
In Figure \ref{fig:pixelHV} (c), position HV $p_3$ and color HV $v_3$ flip at the same time but on different sites, thus the resulting pixel HV $y_3$ flips the two sites. 
Similarly, the Hamming distance is 2 between the new pixel HV $y_3$ and the original pixel HV $y_1$. For the case shown in Figure \ref{fig:pixelHV} (d), position HV $p_4$ and color HV $v_4$ flip at the same site, and the relative Hamming distance is 1.
So far, we have encoded an image with $(r, c)$ size to $r \times c$ HVs.
HDC requires the HVs are pseudo-orthogonal. 
We need to note that the pixel HVs come from position HVs and color HVs, which are pseudo-orthogonal to each other,
and the proof is shown in Lemma \ref{lemma1}.
\newtheorem{lemma}{Lemma}
\begin{lemma} 
\label{lemma1}
Any HVs which will do the element-wise operation in the encoding process, are pseudo-orthogonal.

Proof: 
The basic idea is that although we specify the flip segment, it seems they have a correlation, however, all the correlated HVs will never do an element-wise operation with each other.
From \cite{ge2020classification}, the 2 HVs are orthogonal when their normalized Hamming distance $N(d_h)$ = 0.5. 
Due to the high dimension and random generation, the first row HV $r_0$ and first column HV $c_0$ are pseudo-orthogonal \cite{ge2020classification}. 
This is because 50\% ``1'' and 50\% ``0'' are randomly assigned to the different sites of 2 HVs. 
$r_i$ flips the $x_{row}$ elements based on $r_{i-1}$. 
The flipped $x_{row}$ elements have the same  
probability of distribution of ``1'' and ``0'' with the whole HV.
Thus the number of ``1'' in $r_{i}$ will be very similar to $r_{i-1}$, and thus to $r_0$. This is the same for column HV $c_j$. So, $r_{i}$ and $c_j$ are also pseudo-orthogonal like the $r_{0}$ and $c_0$. Therefore, position HV $p_{(i,j)} = r_i \oplus c_j$ has nearly 50\% ``1'' and 50\% ``0''. 
Color HV $v_k$ is similar to row HV,
with nearly 50\% ``1'' and 50\% ``0''.
Thus, the position HV $p_{(i,j)}$ is also pseudo-orthogonal to the color HV $v_k$, because $N(d_h(p_{(i,j)}, v_k) \approx 0.5)$. 
Note that, there is no
any HV operations between any 2 position HVs, and any 2 color HVs. 
\end{lemma}

\noindent\textbf{\textcircled{\raisebox{-1pt}{4}} Clusterer}

In this work, we use the K-Means algorithm to 
cluster the pixel HVs. 
In the K-means algorithm, the distance function 
is a key part to measure the distances of points to centroids \cite{wu2012cluster}. 
For HDC, Hamming distance and cosine distance are widely used. 
In this work, we use cosine distance to be accommodated with HDC, and the reason will be mentioned later.
The cosine distance of two HVs is defined in Equ \ref{eq:sim}, where $y$ represents the pixel HV, $z$ represents the centroid HV and $d$ is the dimension of HVs.
In classical K-Means, the centroids are chosen randomly. To boost the performance, we choose the pixels with the largest color difference in this work. 
After the first  batch of 
centroids are chosen, the HV distances between all pixels and centroids will be calculated, and each pixel will be clustered to the class of its nearest centroid. After that, all HVs in the same class will be summed to produce the new centroid HV. 
This is the reason we choose the cosine distance. The length of a vector in space will not affect the angle, and the distance will only check the angle.
The process will be conducted iteratively until the preset iteration is achieved, and the pixels have been clustered into different classes.
\begin{equation}
\small
     d_c({\overrightarrow{y}},{\overrightarrow{z}})= 1 - {{\overrightarrow{y}} {\overrightarrow{z}} \over \|{\overrightarrow{y}}\| \|{\overrightarrow{z}}\|} = 1 -\frac{ \sum_{i=1}^{d}{{\overrightarrow{y}}_i{\overrightarrow{z}}_i} }{ \sqrt{\sum_{i=1}^{d}{{\overrightarrow{y}}_i^2}} \sqrt{\sum_{i=1}^{d}{{\overrightarrow{z}}_i^2}} }
    \label{eq:sim}
\end{equation}

%% file: IV_Exp.tex
\section{Experiments}
\label{sec:exp}
This section reports the evaluation results of SegHDC on three nuclei segmentation datasets.
Results show our method outperforms the CNN-based unsupervised segmentation on both segmentation performance and latency.

\noindent\textbf{A. Experimental Setup}

\noindent\textbf{Dataset:}
We employ three segmentation datasets, including BBBC005\cite{ljosa2012annotated} (first 200 images are used), DSB2018\cite{DSB2018} (``stage1\_train'' set is used, cause DSB2018 does not provide the ground truth for the test set), and MoNuSeg\cite{kumar2019multi} (test set is used), to evaluate our method. 

\noindent\textbf{Training setting and baseline:}
The performance of SegHDC is evaluated by comparing it with the 
unsupervised image segmentation method \cite{kim2020unsupervised}. 
The default setting of clustering iteration is 10. The hyperparameters $\alpha$, and $\gamma$ are 
0.2 and 1, while $\beta$ is 
21 on the BBBC005 dataset and 26 on the DSB2018 and MoNuSeg datasets.
The number of clusters 
is set as 2 for BBBC005 and DSB2018 datasets, while 3 is set for the MoNuSeg dataset.
The baseline method runs on the default setting provided by \cite{kim2020unsupervised}.
The metrics of Intersection over Union (IoU) is applied, which is defined as the area of intersection between the predicted segmentation map and the ground truth, divided by the area of union between those two \cite{minaeeimage}. 

\noindent\textbf{Edge device:} To compare the latency of SegHDC and baseline, we employ an edge device: Raspberry PI 4 Model B \cite{RaspberryPI} with 
4 GB memory. The latency is obtained by deploying SegHDC and baseline on the device and testing the processing time for one image.

\noindent\textbf{B . Experimental Results}
\begin{table}[t]
\tabcolsep 4pt
\renewcommand\arraystretch{1.5}
\vspace{-6pt}
  \centering
  \caption{IoU score on 3 datasets}
    \begin{tabular}{|c|c|c|c|c|c|}
    \hline
    \textbf{Dataset} & BL \cite{kim2020unsupervised} & RPos \cite{ge2020classification} & RColor \cite{ge2020classification} & SegHDC & Improvement \\
    \hline
    \textbf{BBBC005} & 0.7490 & 0.0361 & 0.1016 & \textbf{0.9414} & 25.7\%$\uparrow $ \\
    \hline
    \textbf{DSB2018} & 0.6281 & 0.1172 & 0.2352 & \textbf{0.8038} & 28.0\%$\uparrow $ \\
    \hline
    \textbf{MoNuSeg} & 0.5088 & 0.1959 & 0.3832 & \textbf{0.5509} & 8.27\%$\uparrow $ \\
    \hline
    \end{tabular}%
  \label{tab:result}%
\end{table}%

\begin{table}[t]
\vspace{-6pt}
  \centering
  \tabcolsep 4.5pt
\renewcommand\arraystretch{1.5}
  \caption{Result of latency on Raspberry Pi for processing an image in DSB2018 dataset and BBBC dataset}
  \begin{threeparttable}
    \begin{tabular}{|c|c|c|c|c|}
    \hline
          & \textbf{Image Size} & \textbf{IoU Score} & \textbf{Latency on PI} & \textbf{SpeedUp} \\
    \hline
    \textbf{Baseline} & 
    {$256 \times 320 \times 3$ } & 0.7612 & 11453.0s & baseline \\
    \cline{1-1}\cline{3-5}    
    \textbf{SegHDC} &  (DSB2018)     & 0.8275 & 35.8s  & 319.9$\times$ \\
    \hline
    \textbf{Baseline} & 
    {$520 \times 696 \times 1$} & $\times$ \tnote{*} & $\times$ \tnote{*} & baseline  \\
    \cline{1-1}\cline{3-5}    
    \textbf{SegHDC} &  (BBBC005)     & 0.9587 & 178.31s & - \\
    \hline
    \end{tabular}%
    \begin{tablenotes}
        \footnotesize
        \item[$\times ^*$] Out of memory.
      \end{tablenotes}
    \end{threeparttable}
    
  \label{tab:latency}%
\end{table}%

\begin{figure}[t]
\begin{center}
\includegraphics[width=\columnwidth]{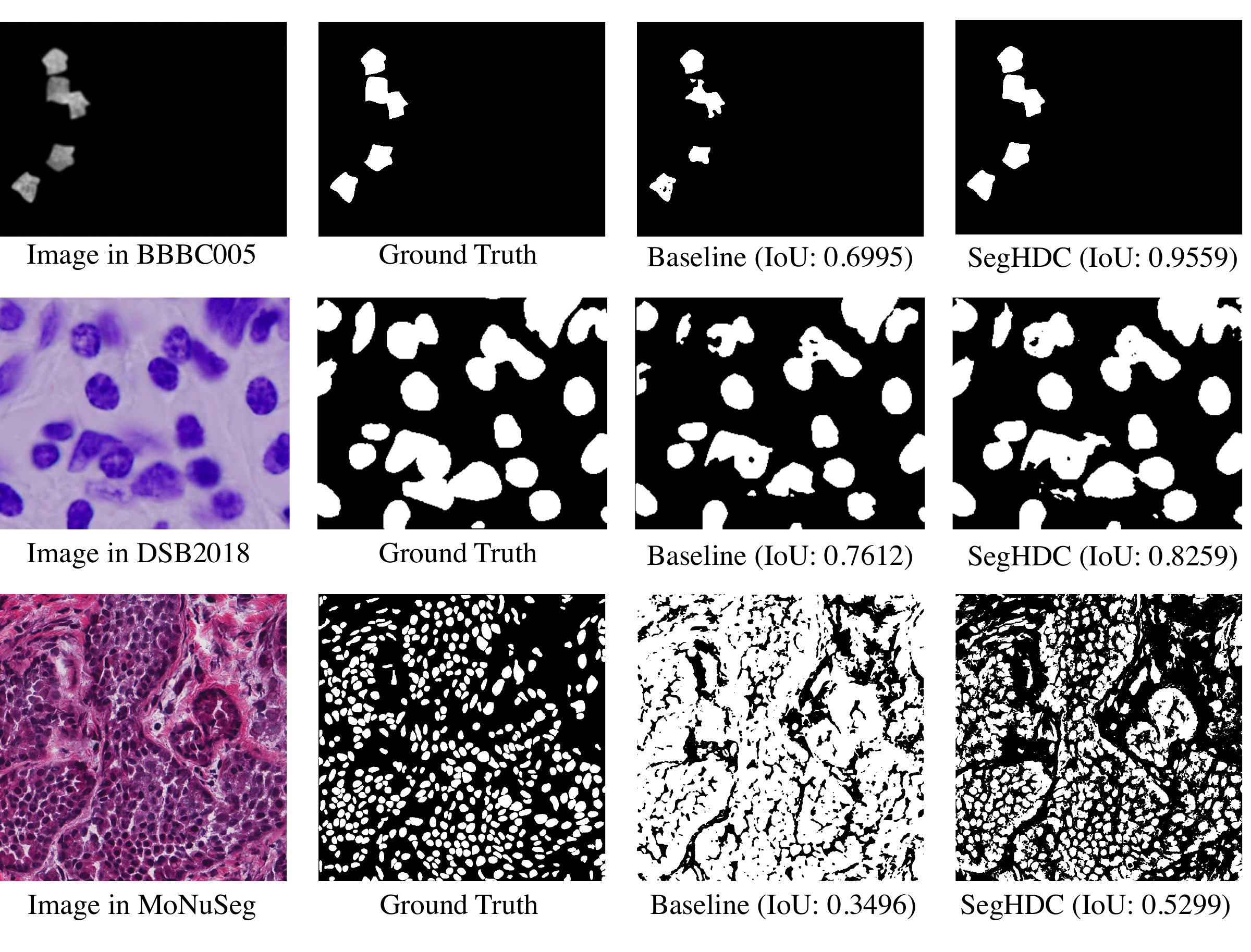}
\vskip -0.1in
\caption{Visualization of prediction masks of an image in BBBC005 dataset, DSB2018 dataset, and MoNuSeg dataset.}
\label{fig:visial2data}
\end{center}
\vskip -0.1in
\end{figure}

\textbf{(1) Our method beats CNN-based baseline}

Table \ref{tab:result} reports the average IoU score on three datasets. Besides the baseline, we also evaluate the encoding method using randomly generated HVs for the position part (denoted as RPos) and for the color part (denoted as RColor) to verify the effectiveness of our encoding method. 

We can have some observations from the results in Table \ref{tab:result}. Overall, SegHDC achieves better results over the baseline on all three datasets. 
Specifically, on the BBBC005 dataset, SegHDC achieves a 0.9414 average IoU score, which is 25.7\% higher than that obtained by the baseline (where the score is 0.749). When it comes to the DSB2018 dataset, our method gets a 0.8038 average IoU score and gains an improvement of 28.0\% compared with 
the score of 0.6281 obtained by the baseline method. On the MoNuSeg dataset, the score of SegHDC surpasses 0.042 over that of the baseline. 
Besides, there is another observation that the results of RPos and RColor are comparable to worse with extremely low IoU scores on these three datasets.
These results have demonstrated the effectiveness of our encoding method.

Table \ref{tab:latency} reported the results of the latency obtained by deploying on the Raspberry Pi for processing an image. An image with size $256 \times 320 \times 3$ from the DSB2018 dataset and an image with size ($520 \times 696 \times 1$) from the BBBC005 dataset are used to test the latency on Raspberry Pi. Specifically,  we use the HV with 800 dimensions, three iterations, and 
$\alpha$ is set as 1 for the image from DSB2018. The baseline method obtains a 0.7612 score, but the latency is over 3 hours as shown in the table. In contrast, SegHDC can obtain a higher IoU score (0.8275) score, meanwhile with a processing time of only 35.8s on the Raspberry Pi, which has achieved $319.9 \times$ speedups than that of the baseline. For the image from BBBC005, the HV is with 2000 dimensions, three iterations are applied, and
$\alpha$ is set as 0.8.
Results show that SegHDC can obtain a 0.9587 IoU score with a latency of 178.31s. However, the baseline approach can not run on the Raspberry Pi (as $\times ^*$ indicated) due to memory limitation.
This set of experiments proves that unsupervised image segmentation is hard to perform on an edge device, where the resources are limited.

Figure \ref{fig:visial2data} demonstrates the sample image, ground truth, prediction mask of baseline and SegHDC, and the IoU score obtained by both approaches. For example, for the sample of BBBC005 in the first line, 
the baseline method can only obtain a 0.6995 IoU score while SegHDC gets a 0.9559 IoU score, where it is obvious to see the difference between the two images in regard to the size and outline of the nuclei. 
When it comes to another sample image DSB2018, the baseline and SegHDC can predict two similar masks (IoU score 0.7612 v.s. 0.8259). However, there is more noise in nuclei obtained by the baseline compared with that obtained by our SegHDC (e.g., some inner noise in nuclei and a small part at the right-up corner in these two images). 
In order to further verify the performance of SegHDC, we have conducted an additional set of experiments on the image in MoNuSeg, where there are much more complex details in the image.
Similar results can be achieved and our proposed SegHDC can predict the mask of most separate nuclei and obtains a higher score of 0.5299. compared with that of the baseline method, where only a rough outline can be predicted with the IoU score of 0.3496.

\textbf{(2) Exploration of SegHDC}

\begin{figure}[t]
\begin{center}
\includegraphics[width=\columnwidth]{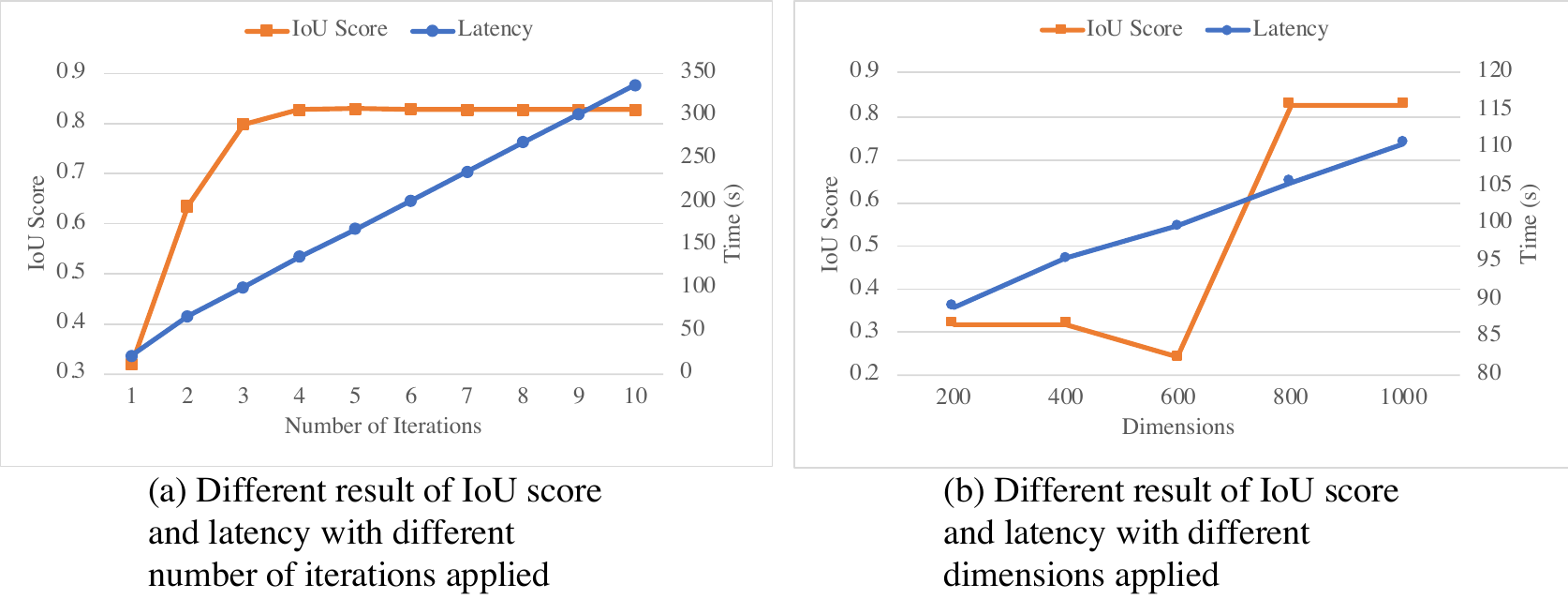}
\vskip -0.1in
\caption{IoU score and latency on Raspberry Pi with different iterations.}
\label{fig:dimAndIter}
\end{center}
\vskip -0.1in
\end{figure}

\begin{figure}[t]
\begin{center}
\includegraphics[width=\columnwidth]{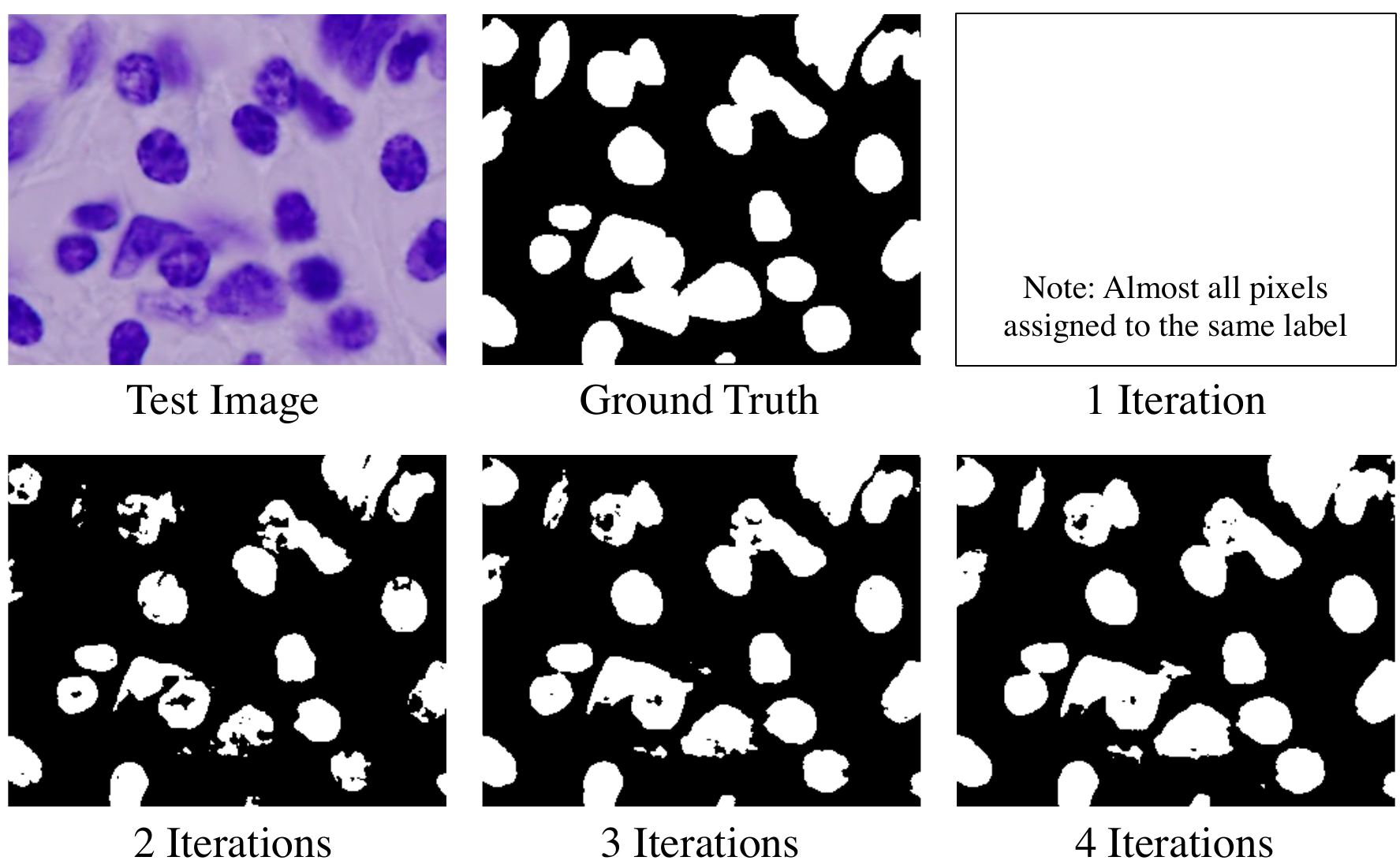}
\vskip -0.1in
\caption{Visualization of prediction masks of an image in DSB2018 dataset during different iterations.}
\label{fig:visial}
\end{center}
\vskip -0.1in
\end{figure}

Except for comparison with the baseline, we also explore how the SegHDC performs with different clustering iterations and dimensions.

By using the sample image in the DSB2018 dataset, 
Figure \ref{fig:dimAndIter} (a) shows the result of the IoU score and the latency on Raspberry Pi with the different number of iterations, specifically when 1 iteration to 10 iterations clustering is applied. Figure \ref{fig:dimAndIter} (b) demonstrates the effect due to the different dimensions applied to this image, where the number of dimensions is set from 200 to 1000. To unify the variables for a fair comparison, the HV dimension is set as 10000 in the experiment shown in Figure \ref{fig:dimAndIter} (a), while the results shown in Figure \ref{fig:dimAndIter} (b) are obtained by processing in 10 iterations.
In both sub-figures, 
the left vertical axis is the IoU score, and the right vertical axis means the latency (s).

There are some observations in this set of experimental results.
As shown in Figure \ref{fig:dimAndIter} (a), the latency time increases from around 20 seconds for 1 iteration to over 300 seconds for 10 iterations. Notated that, after the $4-th$ iteration, SegHDC can obtain a good prediction mask for this image. 
As Figure \ref{fig:dimAndIter} (b) reports, the latency time raises from around 90 seconds for 200-dimension HV to about 110 seconds for 1000-dimension HV. It seems 800 dimensions is a good choice for dealing with this image.
Figure \ref{fig:visial} shows the 
prediction masks of the experiment shown in Figure \ref{fig:dimAndIter} (a). We show the 3-channel test image, ground truth as well as the prediction masks in the first 4 iterations because the later iterations give similar results. 
As shown in the figure, only 1 iteration can not work well,
with more than 2 iterations, it gives a much better result which is close to ground truth.

%% file: V_Conclusion.tex
\section{Conclusion}
\label{sec:conclusion}
To implement the on-device unsupervised image segmentation, we try the first attempt to apply the HDC to perform segmentation. In this paper, we devise a brand new encoding method for both position and color HVs in accordance with Manhattan distance, and propose a segmentation framework, namely SegHDC. 3 nuclei datasets are employed to evaluate the performance of SegHDC. Results show that SegHDC can significantly improve the IoU score with much less latency on Raspberry Pi compared with the CNN-based unsupervised image segmentation approach.